\documentclass[sigconf,nonacm]{acmart}
\AtBeginDocument{%
  }

\usepackage{cleveref}
\usepackage{multirow}
\usepackage{makecell}
\usepackage{algpseudocode}
\usepackage{algorithm}
\usepackage{pifont}
\usepackage{threeparttable}
\usepackage{tikz}
\usetikzlibrary{arrows.meta, positioning, fit, shapes.geometric}

\begin{document}
\newcommand{\errCl}{error class/type}
\newcommand{\justDirty}{Just Dirty}
\newcommand{\dirtyClass}{Dirty \& Type}
\newcommand{\dirtyClean}{Dirty \& Clean}

\newcommand{\qwenLong}{cpatonn/Qwen3-Coder-30B-A3B-Instruct-AWQ-4bit}
\newcommand{\glmLong}{cyankiwi/GLM-4.7-Flash-AWQ-4bit}
\newcommand{\gemmaLong}{RedHatAI/gemma-4-31B-it-FP8-Dynamic}
\newcommand{\Gx}{GX}
\newcommand{\qwen}{Qwen3-Coder}
\newcommand{\glm}{GLM-4.7-Flash}
\newcommand{\gemma}{Gemma-4}
\newcommand{\fname}{\textsc{LeDQeR}}
\newcommand{\cmark}{\ding{51}}
\newcommand{\xmark}{\ding{55}}

\newcommand{\lastname}{\textsf{LastName}}
\newcommand{\firstname}{\textsf{FirstName}}
\newcommand{\emailCol}{\textsf{Email}}
\newcommand{\street}{\textsf{Street}}
\newcommand{\cityCol}{\textsf{City}}
\newcommand{\countryCol}{\textsf{Country}}
\newcommand{\dateBirth}{\textsf{DateOfBirth}}
\newcommand{\countryCode}{\textsf{CountryCode}}
\newcommand{\areaCode}{\textsf{DialingCode}}
\newcommand{\phonenr}{\textsf{PhoneNumber}}
\newcommand{\housenr}{\textsf{HouseNumber}}
\newcommand{\stair}{\textsf{Stairway}}
\newcommand{\door}{\textsf{Door}}
\newcommand{\plz}{\textsf{PostalCode}}

\newcommand{\stepOne}{syntactic filter}
\newcommand{\stepTwo}{correctness filter}
\newcommand{\stepThree}{coverage filter}
\newcommand{\stepFour}{redundancy filter}
\newcommand{\StepOne}{Syntactic filter}
\newcommand{\StepTwo}{Correctness filter}
\newcommand{\StepThree}{Coverage filter}
\newcommand{\StepFour}{Redundancy filter}

\newcommand{\emv}{emv}
\newcommand{\dmv}{dmv}
\newcommand{\con}{con}
\newcommand{\mfv}{mfv}
\newcommand{\ebv}{ebv}
\newcommand{\spm}{spm}
\newcommand{\dov}{dov}
\newcommand{\ifo}{ifo}
\newcommand{\ics}{ics}

\newcommand{\data}[1]{D_{#1}}
\newcommand{\re}[1]{t_{#1}}
\newcommand{\ru}[1]{c_{#1}}
\newcommand{\rules}[1]{C_{#1}}
\newcommand{\rs}[1]{C_{#1}}
\newcommand{\ro}{C}
\newcommand{\rso}{C}
\newcommand{\ruo}{r}
\newcommand{\reo}{t}

\newcommand{\dtrain}{\mathcal{D}}
\newcommand{\candidates}{\mathcal{C}}
\newcommand{\dtest}{\mathcal{D}_{test}}

\newcommand{\rul}{\ensuremath\mathbf{r}}
\newcommand{\rec}{\ensuremath\mathbf{t}}
\newcommand{\err}{\ensuremath\mathbf{e}}

\newcommand{\fsyn}{\ensuremath f_{\textsc{syn}}}
\newcommand{\fcor}{\ensuremath f_{\textsc{cor}}}
\newcommand{\fcov}{\ensuremath f_{\textsc{cov}}}
\newcommand{\fred}{\ensuremath f_{\textsc{red}}}

\newcommand{\rulset}{\ensuremath\mathcal{R}}

\title{Data Quality Rule Generation with LLMs}

\author{Anna-Christina Glock}
\affiliation{%
    \institution{Software Competence Center}
    \city{Hagenberg}
    \country{Austria}
}
\email{anna-christina.glock@scch.at}

\author{Thomas Hütter}
\orcid{0000-0002-7190-6825}
\affiliation{%
    \institution{Software Competence Center}
    \city{Hagenberg}
    \country{Austria}
}
\email{Thomas.Huetter@scch.at}

\author{Johannes Fürnkranz}
\orcid{0000-0002-1207-0159}
\affiliation{
    \institution{Johannes Kepler University}
    \city{Linz}
    \country{Austria}
}
\email{juffi@faw.jku.at}

\author{Wolfram Wöß}
\orcid{0009-0006-8352-0205}
\affiliation{
    \institution{Johannes Kepler University}
    \city{Linz}
    \country{Austria}
}
\email{wolfram.woess@jku.at}

\author{Christine Dominka-Kiss}
\affiliation{%
  \institution{Austrian Post}
  \city{Vienna}
  \country{Austria}
}
\email{christine.dominka-kiss@post.at}

\author{Lisa Ehrlinger}
\orcid{0000-0002-1825-0097}
\affiliation{%
  \institution{Hasso Plattner Institute, University of Potsdam}
  \city{Potsdam}
  \country{Germany}
}
\email{lisa.ehrlinger@hpi.de}

\renewcommand{\shortauthors}{Glock et al.}

\begin{abstract}
The validation of data, such as customer and employee data, is an important task in many organizations.  
Errors in data can have severe consequences. For example, a wrong drug unit in a patient record can lead to life-threatening medication errors, and a missing street number in an address to failed deliveries.
Companies often employ rule-based enterprise data quality (DQ) tools, which allow domain experts to specify rules to validate the data over time. 
While rule-based DQ tools are computationally efficient and provide explainable reports, maintaining a comprehensive rule set manually is challenging, as domain experts often overlook essential rules, especially in complex domains and large data volumes. Hence, closing these gaps remains an open problem in practice.

In this paper, we address the challenge of automated DQ rule generation.
For this, we formalize a generalizable \emph{generate-filter framework} and introduce \fname{}, an \textbf{L}LM-bas\textbf{e}d \textbf{DQ} \textbf{r}ule generation approach. First, a large language model (LLM) generates candidates rules from an observed dirty data tuple for a given rule-based DQ tool syntax. Second, we apply four filter techniques that ensure the (i)~executability, (ii)~correctness, and (iii)~generalizability, and avoid (iv)~redundancy of the generated rules. An extensive experimental evaluation suggests that \fname{} is able to produce effective and compact rule sets for various datasets and error types.

\begin{figure}[t]
    \centering
    \includegraphics[width=\linewidth]{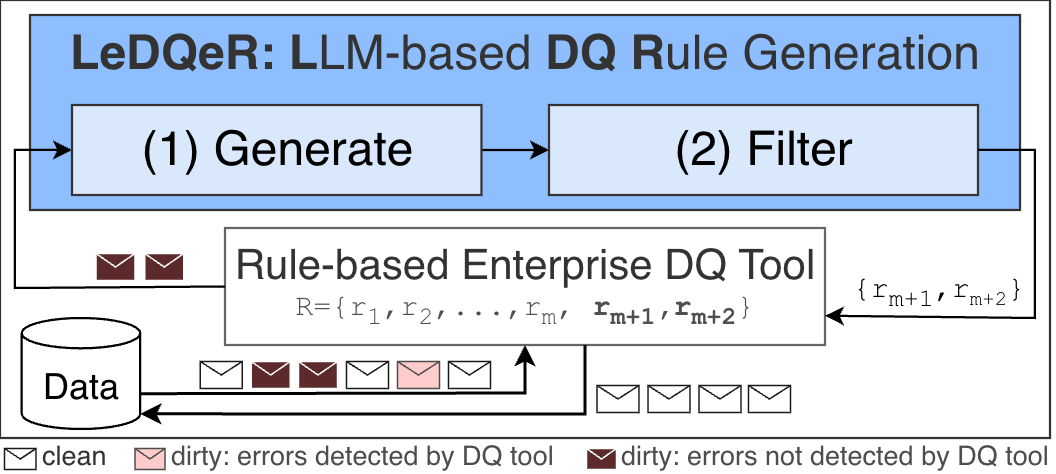}
    \caption{\fname{} extends the rule set of a rule-based enterprise DQ tool. Incoming data is validated against an existing rule set $R = \{r_1, \ldots, r_m\}$. Errors covered by a rule are detected, while errors not covered by any rule pass the validation unnoticed and cause failures in downstream processes. \fname{} closes these gaps: undetected dirty tuples are passed to the generate-filter pipeline, which produces new rules $\{r_{m+1}, r_{m+2}\}$ that extend the existing rule set such that the same error type is detected in the future.}
    \label{fig:ledger_framework}
\end{figure}
\end{abstract}

\ccsdesc[500]{Information systems~Data cleaning}
\ccsdesc[100]{Applied computing~Business rules}
\ccsdesc[100]{Applied computing~Enterprise data management}

\keywords{Data quality, rule generation, rule-based tools, large language models}

\maketitle

\section{Introduction}
\label{sec:introduction}
Data quality (DQ) has long been recognized as a critical prerequisite for effective decision-making in organizations~\cite{WangStrong1996,Redman1998,Pipino2002}. 
\citeauthor{Haug_costOfDQ_2011}~\cite{Haug_costOfDQ_2011} show the impact of poor DQ on organizations that can range from incorrect decision-making and decreased customer satisfaction to life-threatening medication errors in healthcare~\cite{deAndrade_DQHealth_2026}. 
As organizations increasingly deploy artificial intelligence (AI) in practice, the impact of poor DQ becomes even more severe: AI models trained on low-quality data show reduced performance~\cite{mckinsey_DQErrorManu_2023,Mohammed_2025}. Beyond these technical considerations, DQ has also become a regulatory obligation: the EU AI Act mandates that training, validation, and test data of high-risk AI systems must be \emph{relevant}, \emph{representative}, \emph{free of errors}, and \emph{complete}~\cite{EUAIAct2024}. Consequently, even in the era of AI and large language models (LLMs), detecting and correcting data errors in enterprise settings remains an essential requirement.

Machine learning (ML)-based approaches to automate DQ tasks, such as error detection and correction, with minimal expert input have been proposed in research~\cite[e.g.,][]{Mahdavi_raha_2019,HoloClean}. However, these tools lack vendor support and system integration capabilities that are often required by large organizations, and are hence rarely deployed in enterprise settings. 
More recently, also LLMs have been applied directly to detect errors in data~\cite{glock_detecting_2025,Chandru2025,NarayanCOR22,BodensohnBVSB25,yan_adrrLLM_2025}. While effective especially for semantic and context-dependent error types, these approaches require passing the data through the LLM, which is computationally expensive at the scale of enterprise data and typically conflicts with strict data protection requirements, such as the European General Data Protection Regulation (GDPR)~\cite{GDPR2016}, when the data contains personal information. Moreover, \citeauthor{BodensohnBVSB25}~\cite{BodensohnBVSB25} show that the accuracy of LLMs declines sharply on real-world enterprise data, e.g., due to large table sizes and missing internal knowledge.

Consequently, organizations predominantly rely on enterprise DQ tools~\cite{ehrlinger_DQToolSurvey_2022,gartner2021}, which are largely rule-based solutions with some ML and AI support. These tools are computationally efficient, produce deterministic and interpretable results, and are well-established in production. While some enterprise DQ tools recently integrate LLMs to translate natural-language descriptions into rules~\cite{Rehberger2026}, the rule creation itself remains a manual process: a data steward still needs to know and specify which rules are required.

\paragraph{Problem statement.}
However, a rule-based DQ tool is only as good as its rule set. \Cref{fig:ledger_framework} illustrates this setting: incoming data is validated against an existing rule set $R = \{r_1, \ldots, r_m\}$. Data errors that are covered by a rule are detected and can be handled accordingly. Data errors that are not covered by any rule, however, pass the validation unnoticed. As a consequence, the data is classified as clean and propagates into downstream processes, such as ETL pipelines or master data management, where the undetected errors cause processing failures. Consider our industry use case at the Austrian Post: a missing street number or a semantically hard error, such as the confusion of the Austrian cities Lienz and Linz, can lead to failed deliveries. Such failures are particularly costly because they surface late and reoccur until the rule set is extended accordingly. Since manually defining and maintaining DQ rules is complex and time-consuming~\cite{Chiang_2008,ehrlinger_DQToolSurvey_2022,Taleb_BigDQ_2021}, closing these gaps in the rule set remains an open problem in practice.

\paragraph{Our approach.}
In this paper, we solve this problem with \fname{}, a LLM-based framework that iteratively extends the rule set of an enterprise DQ tool whenever an undetected error is identified (cf.~\Cref{fig:ledger_framework}): the dirty tuple that caused a downstream failure is passed to \fname{}, which generates DQ rules that detect the error such that the same error type cannot pass the validation unnoticed again. \fname{} follows a generate-filter approach. In the \emph{generate} step, a LLM automatically generates candidate DQ rules from a dirty tuple and a given DQ rule format. In contrast to traditional rule-learning methods~\cite{Chiang_2008,Yeh_2010}, which struggle to capture semantic context and frequently require labeled data, LLMs leverage pre-trained knowledge and semantic reasoning capabilities to generate DQ rules in code-like syntax (e.g., SQL or Python) from minimal input~\cite{Yadav_zeroShotCode_2025,Raghav_tabQuestion_2025}. In the \emph{filter} step, the candidate rules are progressively refined with respect to (i)~executability, (ii)~correctness, (iii)~generalizability, and (iv)~redundancy, producing a minimal and precise rule set that is ready for operational deployment.

\paragraph{Contributions}
In summary, this paper makes the following contributions to automate DQ rule generation in enterprise settings:
\begin{itemize}
    \item We formalize the problem of DQ rule generation and introduce a generalizable generate-filter framework.
    \item We propose \fname{}, a LLM-based implementation of the framework that generates DQ rules from observed dirty tuples and incrementally extends an existing rule set.
    \item We introduce a suite of four filter techniques that ensure the (i)~executability, (ii)~correctness, (iii)~generalizability, and (iv)~redundancy of the new rules. 
    \item We comprehensively evaluate \fname{} on five datasets from diverse domains with up to nine different error types.
\end{itemize}

\paragraph{Results}
Our evaluation shows that \fname{} enables the generation of executable rules that generalize well to unseen data without producing false positives. In particular, our experiments suggest that LLMs are promising for generating DQ rules and show the effectiveness of our filters. We also observed that the ability of the generated rules to detect a dirty tuple is influenced by the error type and the amount of error context provided in the prompt.

\paragraph{Outline}
We first introduce necessary background on data errors and rule-based DQ tools in \Cref{sec:preliminaries}, where we also present the running example used throughout this paper.
In \Cref{sec:approach}, we first formalize the problem of DQ rule generation and subsequently propose \fname{}, an LLM-based framework for DQ rule generation that follows our generate-filter formalization. We evaluate \fname{} in \Cref{sec:experiments} and discuss related work in \Cref{sec:related-work}. \Cref{sec:conclusion} concludes this paper with an outlook on future work.

\section{Data Errors and Rule-based DQ Measurement}
\label{sec:preliminaries}
In this section, we introduce the necessary background on data error types in \Cref{subsec:errType} and rule-based DQ tools in \Cref{sec:runningSample}. 
Throughout this section, we will also introduce a running example that is derived from our industry use case at the Austrian Post: a personal contact information (PCI) dataset (cf.~\Cref{tab:runingSample_exp}).
This example serves as a walkthrough to explain our generate-filter framework in \Cref{sec:approach} and its experimental evaluation in \Cref{sec:experiments}.

\subsection{Data Quality and Error Types}
\label{subsec:errType}
\Cref{tab:runingSample_exp} shows our running example: a subset of a personal contact information (PCI) dataset from our industry use case at the Austrian Post. Each tuple describes a person and their address with the columns \street{}, \firstname{}, \plz{}, and \cityCol{}. The example is a reduced version of the full \texttt{PCI} dataset with $14$ columns, which we use in our experimental evaluation (cf.~\Cref{sec:experiments}).

Data quality is per definition context-dependent, as it is commonly defined as ``fitness for use''~\cite{WangStrong1996}, which means that the data must be fit for the specific purpose it is used for. Consequently, DQ measurement must reflect domain-specific semantics and usage expectations.

\begin{example}
\label{ex:data_usecase}
The use case of our running example  introduces the following requirements on the data: (1)~street names must contain actual street names and no placeholder values such as ``Unknown'', (2)~a four-digit numerical representation like ``1220'' is required for the postal code, even though a leading ``A'' is commonly added in Austria (e.g., ``A-1220''), and (3)~the city solely consists of the city name, with no subareas being allowed. Values that violate these requirements are data errors and are highlighted in color in \Cref{tab:runingSample_exp}.
\end{example}

\begin{table}[t]
  \caption{Subset of the PCI dataset as running example.}
  \label{tab:runingSample_exp}
  \begin{tabular}{rllll}
    \toprule
    ID & \street & \firstname & \plz & \cityCol\\
    \midrule
    1 & \textcolor{blue!80}{Unknown} & Ina-Maria & \textcolor{red!80}{A-1220} & Wien \\
    2 & Brahmsplatz & Hans & 1040 & \textcolor{orange}{Wien, Wieden} \\
    3 & Berggasse & Anna & \textcolor{red!80}{A-4020} & Linz \\
    4 & Feldgasse & Anna-Maria & \textcolor{red!80}{A6020} & Innsbruck \\
    5 & \textcolor{blue!80}{Unknown} & Maria & 8010 & Graz \\
  \bottomrule
   \addlinespace[5pt]
   \multicolumn{5}{l}{Type of error: \hspace{0.1cm}\hspace{0cm}\textcolor{orange!80}{$\blacksquare$} Embedded Value}\\
   \multicolumn{5}{l}{ \hspace{1.96cm}\textcolor{red!60}{$\blacksquare$} \textsf{Incorrect Format} \hspace{0.1cm}\textcolor{blue!60}{$\blacksquare$} Disguised Missing Value}\\
\end{tabular}
\end{table}

As illustrated in Example~\ref{ex:data_usecase}, data errors are observable effects of poor data quality and are inherently context- and usage-specific. This has led to the development of numerous data-error taxonomies~\cite{RahmDo2000,Müller2003,Kim2003,Oliveira2005,Josko2016,Ilyas2022,bhadauria2026}.  Based on these taxonomies and feedback from domain experts, we selected nine error types that cover a broad range of DQ issues that appear in single values or tuples for evaluating our framework:

\begin{description}
    \item[Explicit missing values (\emv)] are values explicitly set to null. This is a common error type~\cite{pearson_dmv_2006} and well-known in literature~\cite{RahmDo2000,Oliveira2005,Kim2003,Müller2003,Ilyas2022,jung2025,bhadauria2026}.  
    
    \item[Disguised missing values (\dmv)] are semantically missing (like \emv) but contain placeholder values. For example, John Doe as \firstname{} and \lastname{} or 01.01.1900 as \dateBirth{}. Detecting this error type is challenging because they might be syntactically valid but semantically incorrect~\cite{pearson_dmv_2006,bhadauria2026}.
    
    \item[Contradictions (\con)] are
    mutually inconsistent values among different values of a tuple~\cite{Müller2003}. 
    For example, both the postal code and the city may be syntactically correct, but do not match the same entity.
    Even though literature sometimes describes such errors under the umbrella of ``functional dependency violations''~\cite{Oliveira2005,Josko2016}, practical scenarios can be more complex.
    For example, no functional dependency holds between city and postal code. In other words, one postal code can refer to multiple cities (e.g., 4232 is the postal code for $9$ municipalities in Austria), whereas one city can have multiple postal codes (e.g., Vienna has $23$ postal codes). Therefore, we use the more general term ``contradiction'' introduced by \citeauthor{Müller2003}~\cite{Müller2003} in this paper.
    
    \item[Misfielded values (\mfv)] are correct values that are switched and therefore end up in the wrong field/column. For example, a value of a \firstname{} is stored in the respective \lastname{} of the same tuple and vice versa. This error type  has already been described by~\cite{Kim2003,RahmDo2000,jung2025,bhadauria2026}.  
    
    \item[Embedded values (\ebv)] are values that contain additional, unwanted information~\cite{Oliveira2005,RahmDo2000,bhadauria2026}. For example, the postal code ``1220, Donaustadt'' also contains the district name.
    
    \item[Spelling mistakes (\spm)] are incorrect values resulting from added, deleted, or exchanged characters/numbers~\cite{RahmDo2000,Oliveira2005,Kim2003,Josko2016,jung2025,bhadauria2026}. Common examples include typos, such as missing an ''n'' in ''Viena''.
    
    \item[Domain violations (\dov)] refer to values that are outside of an expected range or violates business constraints~\cite{Oliveira2005,Josko2016,Müller2003}. For example, a negative \phonenr{} or the violation of a business rule that enforces the country column to contain the full name of the country.
    
    \item[Incorrect formats (\ifo)] may appear for various columns and data formats. For example, the date might follow different standards ('MM/DD/YYYY' vs. 'DD/MM/YYYY'). This error type is listed as a syntax violation in~\cite{Oliveira2005,bhadauria2026}, as a domain format error in~\cite{Müller2003}, and as a format error in~\cite{jung2025}.
    
    \item[Incorrect character sets (\ics)] are used in the data. For example, the Cyrillic character set was used to write the address instead of the Latin character set. 
    This error type was also introduced by \citeauthor{Kim2003}~\cite{Kim2003}, who focus on wrong encodings (e.g., ASCII vs. EBCDIC) and also mentioned as incorrect encoding in~\cite{jung2025,bhadauria2026}.
\end{description}
While not exhaustive, the selected error types capture the major classes of DQ issues described in literature and encountered by practitioners.

\begin{example}
\label{ex:error_classification}
Each tuple in \Cref{tab:runingSample_exp} is identifiable via column ID and contains at least one dirty value, which can be classified as one of the error types: the errors in \street{} are of type \dmv{}, as two tuples contain the value \textit{Unknown}, which indicates a missing street name without actually being a missing value. \firstname{} has no errors. Three values in \plz{} are of error type \ifo{}, since the correct format is the four-digit postal code as shown in tuples 2 and 5. Tuple 2 violates the requirement that the city is limited to the city name and is hence of type \ebv{}.
\end{example}

\subsection{Rule-based Data Quality Measurement}
\label{sec:runningSample}
Rule-based DQ measurement~\cite{Loshin2002} is a foundational approach to measuring data quality, relying on an explicitly defined rule set that captures the expected data properties. These rules formalize domain knowledge and business constraints (e.g., allowable value ranges, referential integrity conditions, or pattern-based formats) and are systematically evaluated against datasets to detect violations. 

Several practical data quality tools and frameworks support the rule‑based approach~\cite{ehrlinger_DQToolSurvey_2022}. Prominent open‑source systems include Great Expectations\footnote{\url{https://github.com/great-expectations/great_expectations}}(\Gx), Amazon Deequ\footnote{\url{https://github.com/awslabs/deequ}}, and Apache Griffin\footnote{\url{https://griffin.apache.org/}}. In the commercial domain, platforms such as Informatica\footnote{\url{https://www.informatica.com/products/data-quality.html}}, Talend Data Quality\footnote{\url{https://www.talend.com/products/data-quality/}}, and Ataccama\footnote{\url{https://www.ataccama.com/platform/data-quality}} offer support for rule-based data quality evaluation. Despite differences in implementation details, these tools share a common architecture in which explicitly defined rules are evaluated against data to identify violations that subsequently provide data quality indicators. 
In the remainder of this paper, we focus on \Gx{} as a representative rule‑based data quality framework due to its widespread adoption and open‑source nature. A list of expectations (i.e., rules) in \Gx{} syntax is shown in Example~\ref{ex:gx_rules}.

\begin{example}
\label{ex:gx_rules}
    Based on the data in Table~\ref{tab:runingSample_exp}, a set of hypothetically generated, possibly incorrect rules\
    is given below:
    \begin{enumerate}
        \item gx.expectations.ExpectColumnValuesToNotMatchRegex( \\
        column='\plz', regex=\verb|r'A-[0-9]*'|)
        \item gx.expectations.ExpectColumnValuesToMatchRegex(\\
        column='\street', regex=\verb|r'^Unknown$'|)
        \item gx.expectations.ExpectColumnValuesToNotMatchRegex( \\
        column='\firstname', regex=\verb|r'^[A-Z][a-z\-]*$'|)
        \item gx.expectations.ColumnNotMatchRegex( \\
        column='\street', regex=\verb|r'[A-Z][a-z]*'|)
    \end{enumerate}

Each rule specifies a pattern with a regular expression, where the operation defines whether matching values are considered correct or incorrect.
Rule (1) states that postal codes starting with 'A-' are incorrect, which correctly identifies the entries marked in red in Table~\ref{tab:runingSample_exp}. 
Rule~(2) is an example for an incorrect rule because it expects street names to match the value ``Unknown'', thereby erroneously flagging the valid street names in tuples 2--4 while accepting the actual errors. 
Rule~(3) states that first names that consist of a capitalized letter followed by lower-case letters, possibly separated by hyphens, are incorrect, which would erroneously mark entries 2, 3, and 4 of the first name column as errors. 
Rule~(4) syntactically violates the \Gx{} format and is not executable at all. 
Note that these hypothetical rules are intentionally imperfect and illustrate typical problems that occur during rule generation. In \Cref{sec:approach}, we show how our filters remove such rules.
\end{example}

While rule-based DQ frameworks provide deterministic and transparent data validation, there are some notable limitations: First, data stewards often lack complete insight into newly emerging or fast evolving data domains, which makes it difficult to define comprehensive rule sets upfront and maintaining hundreds of rules over time. Second, defining rules for semantic error types (e.g., ``empty'' refers to a missing value) is more difficult than for syntactic error types (e.g., Austrian postal codes must consists of four digits). 
In summary, the manual generation of suitable DQ rules remains a huge and practical challenge in enterprise settings.

\section{LLM-based Data Quality Rule Generation}
\label{sec:approach}

We introduce a LLM-based approach for data quality rule generation that addresses the shortcomings of manual approaches. 
In this section, we formalize the problem, introduce a general generate-filter framework for data quality rule generation, and then present our LLM-based approach to instantiate the framework.

\subsection{Problem Formalization}
\label{sec:approach-form}
Data quality rule generation aims at automatically deriving a set of DQ rules $\rulset = \{r_1, r_2, \ldots, r_m\}$ for a given dataset $\dtrain = \{t_1, t_2, \ldots, t_n\}$ of $n$ tuples and a rule-based DQ tool supporting rule syntax $\mathcal{S}$. The quality of the generated rules depends on their executability (i.e., syntactic correctness) and their ability to identify the corresponding error types in the data.

\begin{figure}[H]
    \centering
    \begin{tikzpicture}[
        node distance=2.5cm,
        block/.style={draw, rectangle, minimum width=2.6cm, minimum height=1.2cm},
        arrow/.style={->, thick}
    ]
    
    \node (in) {};
    \node[block, right=1cm of in] (gen) {Generate};
    \node[block, right=1cm of gen] (filter) {Filter};
    \node (out) [right=1cm of filter] {};
    
    \draw[arrow] (in) -- node[above] {$\dtrain$} (gen.west);
    \draw[arrow] (gen.east) -- node[above] {$\candidates$} (filter.west);
    \draw[arrow] (filter.east) -- node[above] {$\rulset$} (out);
    
    \end{tikzpicture}
    \caption{Generate-filter framework for DQ rule generation.}
    \label{fig:generate_filter_framework}
\end{figure}
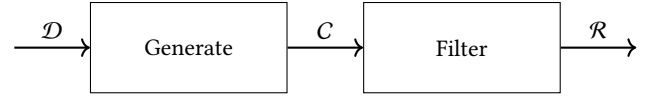

To foster the discussion of existing~\cite{Xie_DQRLLMmed_2025,Schneider_LLMBagOfRules_2025}, our proposed, and future approaches, we introduce a generalized framework for data quality rule generation, called \emph{generate-filter framework} (cf.~\Cref{fig:generate_filter_framework}). The framework produces a rule set $\rulset$ in two steps:

\begin{enumerate}
\item \emph{Rule Generation:} Given an input dataset $\dtrain$, the objective is to generate an initial set of candidate DQ rules $\candidates$. 
\item \emph{Rule Filtering:} Since the generated rule candidates $\candidates$ may include erroneous, redundant, or overlapping rules, a subsequent filtering phase is applied to identify and remove low-quality rules. Thereby, the overall correctness and coverage of the resulting rule set $\rulset$ is improved.
\end{enumerate}

\begin{example}

Consider dataset $\dtrain$ from Example~\ref{ex:data_usecase}. In the generate step, a set of $|\candidates| = 4$ candidate rules for the DQ tool \Gx~is generated (cf. Example~\ref{ex:gx_rules}). Assume that in the filtering step, the syntactical correctness of the rules is verified. 
Consequently, $\rulset$ contains three rules (1-3), while rule (4) is discarded because it does not conform to \Gx's syntax. 
\end{example}

\begin{figure*}
    \centering

    \includegraphics[width=1\linewidth]{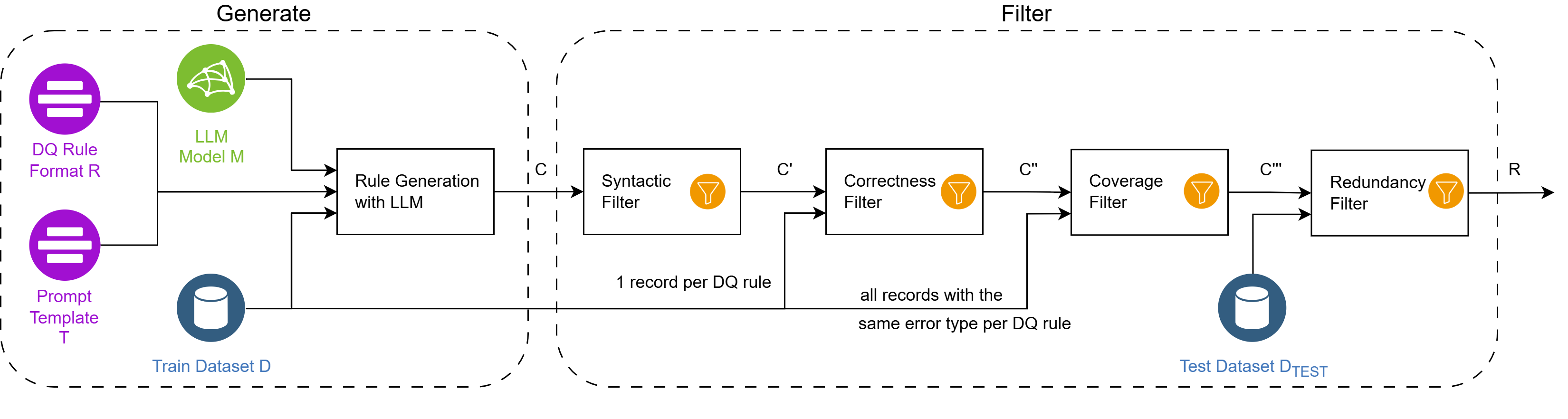}
    \caption{Architecture of \fname{}: an LLM generates candidate rules $\candidates$, which are progressively refined through four filters $\candidates \supseteq \candidates' \supseteq \candidates'' \supseteq \candidates'''$ until a final rule set $\rulset$ is returned.}
    \label{fig:eval_pip}
\end{figure*}

\subsection{\fname: An LLM-based Generate-Filter Framework to Generate DQ Rules}
\label{sec:approach-ledger}
We introduce \fname{}, a LLM-based instantiation of the generate-filter framework. 
\Cref{fig:eval_pip} shows the architecture of \fname{}: First, we generate a candidate DQ rule set using a LLM-based technique. Second, we propose a set of four complementary filters that progressively refine the candidate rule set $\candidates \supseteq \candidates' \supseteq \candidates'' \supseteq \candidates''' \supseteq \rulset$.  

\renewcommand{\algorithmicrequire}{\textbf{Input:}}
\renewcommand{\algorithmicensure}{\textbf{Output:}}
\begin{algorithm}[b]
\caption{DQ Rule Generation}
\label{alg:rule_generation}
\begin{algorithmic}[1]
\Require $\dtrain$ (Dirty Dataset), $\mathcal{M}$ (LLM Model), $\mathcal{P}$ (Prompt Template), $\mathcal{S}$ (Rule Syntax), $Desc$ (Dataset Description)
\Ensure $\candidates$ (Candidate set of DQ Rules)
\Procedure{generateRules}{$\dtrain, \mathcal{M}, \mathcal{P}, \mathcal{S},Desc$}
    \State $\candidates \gets \emptyset$
    \For{$\rec_i \in \dtrain$}
        \State $prompt \gets \text{constructPrompt}(\mathcal{P}, \rec_i, \mathcal{S}, Desc)$
        \State $response \gets \text{queryLLM}(prompt, \mathcal{M})$
        \State $\candidates_i \gets \text{parseResponse}(response)$
        \State $\candidates \gets \candidates \cup \candidates_i$
    \EndFor
    \State \Return $\candidates$
\EndProcedure
\end{algorithmic}
\end{algorithm}

\subsubsection{LLM-based DQ Rule Generation}
\label{subsec:exp_ruleGen}
Instead of defining the set of DQ rules by hand, as it is typically expected by state-of-the-art DQ tools (cf.~\autoref{sec:preliminaries}), we present an automated LLM-based approach to generate DQ rules. 
Algorithm~\ref{alg:rule_generation} describes the generation process, which requires the following inputs: a (dirty) dataset $\dtrain$, a LLM model $\mathcal{M}$, a prompt template $\mathcal{P}$, a DQ rule syntax $\mathcal{S}$, and 
a textual description $Desc$ of the semantic context of the dataset
(e.g., ``\textit{The data contain information about the scheduled and actual departure and arrival times of airplanes for a flight dataset.}''). 
Each tuple $\rec_i \in \dtrain$ contains at least one error.  For each tuple $\rec_i$, a prompt template $\mathcal{P}$ is populated with the dirty record, a guideline for the DQ rule syntax $\mathcal{S}$, and an overall dataset description. 
For this prompt, the LLM 
generates a set of candidate rules $\candidates_i$ for tuple $\rec_i$. 
The number of generated rules $|\candidates_i|$ may differ from the actual error count of the dirty record, as the LLM is unaware of the exact number of errors and may generate multiple rules or none for a specific error. The global candidate set $\candidates$ is defined as the union of all individual rule sets $\candidates_i$ generated for each tuple $\rec_i$ with $1 \leq i \leq n$, such that $\candidates = \bigcup_{i=1}^{n} \candidates_i$.

\paragraph{Prompt construction.}
The prompt is a central part of the generation process and consists of the following four sections: the prompt template $\mathcal{P}$ (containing the initial framing and tasks), the input data $\rec_i$, the rule syntax $\mathcal{S}$, and the dataset description $Desc$.
For each tuple $\rec_i$, the function $\text{constructPrompt}(\mathcal{P}, \rec_i, \mathcal{S}, Desc)$ instantiates the prompt template into a concrete prompt (cf.~Algorithm~\ref{alg:rule_generation}, line~4).

The \emph{initial framing} of the prompt contains overall information such as the LLM's role (e.g., ``\textit{You are a \Gx{} rule expert.}'') and the overall task (e.g., ``\textit{Generate \Gx{} rules to prevent an error in the data sample}). 

We support and evaluate three variants of providing the \emph{input data} $\rec_i$ with increasing difficulty: 
\begin{enumerate}
    \item \emph{\dirtyClean:} The LLM is provided with clean and dirty data. While this scenario is uncommon in practice, it provides the most information to the LLM and serves as an upper bound for our evaluation.
    \item \emph{\dirtyClass:} In addition to dirty data, the LLM also receives information about the error types present in the columns of a record. 
    This scenario is applicable in case the applied error detection approach returns additional error type information rather than a binary assessment. 
    \item \emph{\justDirty:} Only dirty data is provided to the LLM. This variant is the most realistic scenario in practice, where no additional data and ground truth are available. Compared to the other input data variants, the LLM needs to detect the error without additional information before generating a rule. 
\end{enumerate}

The \emph{rule} section specifies the types of rules the LLM should generate, including information about the DQ tool: 
(1) A clear description of the application scenario, e.g., focus on generating a single rule set without additional information.
(2) Details on the targeted DQ tool, e.g., \Gx{} version numbers and rule syntax $\mathcal{S}$ examples. 
This includes information on generating custom rule functions as well as a list of existing \Gx{} rule functions with (e.g., ''\textit{ExpectColumnValuesToNotMatchRegex(column: str, regex: str)}'') and without parameters (e.g., ''\textit{ExpectColumnValuesToNotMatchRegex}''). 
Otherwise, there is a high chance to generate incorrect rules.

In the last prompt section, we guide the LLM by defining four consecutive sub-tasks to achieve the overall goal:
\begin{enumerate}
    \item detect the error in the provided data,
    \item generate rules to detect the identified errors,    
    \item check if the rules are general and fit the given syntax, and 
    \item return the rules in a specific format (in our case JSON).
\end{enumerate}

\subsubsection{Filtering pipeline}
\label{subsec:filtering_pipeline}

After generating candidate rules $\candidates$ in the first step of the framework, filters are applied to improve the overall quality of $\candidates$. 
In \fname{}, we implement a set of four filters that are sequently applied to progressively refine the rules and produce a final rule set $\candidates \supseteq \candidates' \supseteq \candidates'' \supseteq \candidates''' \supseteq \rulset$.

\paragraph{Syntactic filter}
\label{subsubsec:syntactic_filter}
The syntactic filter verifies whether the generated rules $\candidates$ are executable and match the given DQ rule syntax $\mathcal{S}$. The result is a reduced candidate rule set $\candidates' \subseteq \candidates$ that contains only syntactically correct (i.e., executable) rules. No additional information, such as labeled data, is required for this filter. 
\begin{example}
 For Example~\ref{ex:gx_rules}, this means that the syntactical correctness rate is $0.75$ since the fourth rule contains an operation that is not supported in \Gx{} (\texttt{ColumnNotMatchRegex}) and hence only $3$ out of $4$ rules are syntactically correct. 
\end{example}

\paragraph{Correctness filter}
\label{subsubsec:rule_verification}
The correctness filter removes rules that fail to filter the tuple from which they were generated. Therefore, we re-apply and verify each generated rule $\rul \in \candidates'$ on the tuple that has triggered its generation. A rule either recognizes the error (\emph{true positive TP}), fails to recognize the error (\emph{false negative FN}), or mistakenly detects an error in a clean value (\emph{false positive FP}). The filter removes all rules that do not lead to a TP from $\candidates'$ leading to a reduced set of rules $\candidates'' \subseteq \candidates'$.

\begin{example}
Applying this filter to the rules in our Example~\ref{ex:gx_rules} with the data from \Cref{tab:runingSample_exp}
yields an F1 score of 0.5. Specifically, the first rule correctly detects that the \plz{} is wrong (TP). 
The \street{} is wrong but not detected by any rule (FN). Rule three incorrectly flags the clean \firstname{} as wrong, resulting in a false positive (FP). 
\end{example}

\paragraph{Coverage filter}
\label{subsubsec:coverage_filter}
To identify rules $\rul \in \candidates''$ that do not generalize well beyond the original error instance within the observed input tuple $\rec_i$, the coverage filter tests each rule against all tuples that share the same error type in the training dataset $\mathcal{D}$. Rules are removed if they achieve a recall above a predefined threshold across these variants, resulting in $\candidates''' \subseteq \candidates''$. While precision was addressed in correctness filter, the coverage filter focuses exclusively on recall. A high recall indicates strong generalizability across various variants of the same error type.
However, a low recall does not necessarily imply that a rule is ineffective, as some errors may require highly specific rules to avoid confusion with correct values. Unfortunately, such rules may exhibit low recall and could be filtered out depending on the chosen threshold.
In the current implementation of \fname{}, we do not set a minimum recall threshold to prioritize maximum coverage and avoid the risk of filtering highly specific rules. 

\begin{example}
The remaining rule in $\ro''$ is applied on tuples 1, 3 and 4 (\Cref{tab:runingSample_exp}). Tuple 2 has a different error and thus will not be used. 
Tuple 5 shares one error column and error type with tuple 1. However, the LLM did not generate a rule to detect this error successfully in tuple 1, thus the tuple is not selected.
As only rule 1 passed the two previous filters, we consider only this rule for the coverage filter. The recall of 0.66 is calculated from 
\begin{itemize}
    \item TP (2): tuple 1 (\plz), tuple 3 (\plz)
    \item FP (0): 
    \item FN (1): tuple 4 (\plz)
\end{itemize}

This means the generalizability of the rule is not optimal, as it captures only a subset of error variants. 
While it successfully detects the dirty tuple 3 (with the \plz{} format "A-<4 digits>"), it fails to identify another dirty tuple (with the \plz{} format "A<4 digits>"). As we do not set a minimum recall threshold, $\ro'''$ consists of rule 1.

\end{example}

\paragraph{Redundancy filter}
\label{subsubsec:redundancy_filter}
The redundancy filter removes duplicated and overlapping rules by using a greedy covering algorithm to reduce the learned rules to a minimal rule set that collectively covers all errors of an error type~\cite{jf:AI-Review}. We therefore use a separate test dataset $\dtest$ to estimate precision and recall. In practice, $\dtest$ can be a curated dataset used to validate rule performance. While our evaluation employs large-scale datasets to demonstrate scalability, smaller curated sets are sufficient for practical application. We iteratively select rules with the highest recall among those achieving precision $1$, mark their covered tuples, and repeat until all tuples are covered, resulting in the final rule set $\rulset \subseteq \candidates'''$. Since false positives are costly in practice, precision is the primary filter metric.
Unlike the previous filters, the redundancy filter removes overlapping rules in the $\candidates'''$ across an unseen $\dtest$. Consequently, its behavior cannot be meaningfully demonstrated using the small-scale \Cref{tab:runingSample_exp}. 
We therefore defer its illustration to the experimental evaluation in \Cref{sec:eval-redundancy}.

\section{Experiments}
\label{sec:experiments}

In this section, we experimentally evaluate \fname{} by analyzing each step of the generate-filter framework: the rule generation and the \stepOne{} in \Cref{sec:eval-syntactic}, the \stepTwo{} in \Cref{sec:eval-correctness}, the \stepThree{} in \Cref{sec:eval-coverage}, and the \stepFour{} in \Cref{sec:eval-redundancy}. The experimental setup is described in \Cref{subsec:impl}. As each step targets a different aspect of rule quality, the datasets, evaluation metrics, and number of rules vary by experiment.

\subsection{Experimental Setup}
\label{subsec:impl}
All experiments were implemented in Python and are available on GitHub\footnote{https://github.com/Anna-Christina-Glock/dq\_rule\_generation}.
We ran the experiments on our high-performance computing platform, which features a single AMD EPYC 7643 (48-core, 2.3 GHz) CPU, four NVIDIA A100 GPUs (each with 160GB high-bandwidth memory), and 1TB RAM.

\subsubsection{Data quality tool}
\label{subsec:dqTool}
In our experiments, we create rules for the open source DQ tool Great Expectations(\Gx), which  enables users to create rules (``expectations'' in \Gx{} terminology) to check whether the validated data contains errors (see Example~\ref{ex:gx_rules} for a syntax example). \Gx{} provides multiple predefined rules that can be customized.
We have chosen \Gx{} in version 1.9.1 (i.e., the latest stable version) for our experiments, since it is freely available, based on Python, has a large and active open-source community, as well as active partnerships with major industry companies such as Snowflake and Databricks~\cite{gx_partners}. 

\subsubsection{Datasets}
We used five datasets for our evaluation, summarized in \Cref{tab:datasets}. The \texttt{beers}, \texttt{flights}, \texttt{hospital}, and \texttt{tax} datasets are adapted from Mahdavi et al.~\cite{Mahdavi_raha_2019}. The \texttt{PCI} (personal contact information) dataset is synthetically generated from real Austrian addresses~\cite{glock_detecting_2025}. Compared to the original version in~\cite{glock_detecting_2025}, we polluted the \texttt{PCI} dataset with additional error types for this study\footnote{https://github.com/Anna-Christina-Glock/pci-llm-toolkit/tree/main/data-generator}.

\begin{table}[H]
    \centering
    \caption{Dataset characteristics used in our experiments.}
    \label{tab:datasets}
    \begin{tabular}{lrrr}
        \toprule
        Dataset & \# Columns & \# Dirty Tuples & \# Error Types \\
        \midrule
        \texttt{Beers}    & 10 & 2{,}410 & 1 \\
        \texttt{Flights}  &  7 & 1{,}904 & 3 \\
        \texttt{Hospital} & 19 &    407  & 1 \\
        \texttt{Tax}      & 15 & 6{,}000 & 2 \\
        \texttt{PCI}      & 14 & 6{,}000 & 9 \\
        \bottomrule
    \end{tabular}
\end{table}

 \begin{table*}[t]
   \caption{\StepOne: Number of generated rules and the corresponding percentage of executable rules across three LLMs, five $\dtrain$, and three input variants (\dirtyClean{}, \dirtyClass{}, \justDirty{}), and two $\mathcal{S}$ variations (with vs. without parameter information).}
   \begin{tabular}{llrrrrrr}
\toprule
 & & \multicolumn{2}{c}{GLM-4.7} & \multicolumn{2}{c}{Gemma-4} & \multicolumn{2}{c}{Qwen3-Coder} \\
\cmidrule(lr){3-4} \cmidrule(lr){5-6} \cmidrule(lr){7-8}
 & & no Parameter & Parameter & no Parameter & Parameter & no Parameter & Parameter \\
\midrule
\midrule
\multicolumn{8}{l}{\textbf{Dirty \& Clean}} \\
\midrule
\multirow{2}{*}{\texttt{Beers}} & No. Rules & 2{,}534 & 2{,}569 & 2{,}984 & 3{,}119 & 3{,}223 & 3{,}137 \\
 & \% Executable & 0.28 & 0.84 & 0.99 & 1.00 & 0.98 & 0.94 \\
\multirow{2}{*}{\texttt{Flights}} & No. Rules & 2{,}002 & 2{,}748 & 3{,}154 & 902 & 7{,}860 & 7{,}804 \\
 & \% Executable & 0.45 & 0.77 & 0.95 & 0.93 & 0.98 & 0.94 \\
\multirow{2}{*}{\texttt{Hospital}} & No. Rules & 468 & 502 & 476 & 484 & 530 & 621 \\
 & \% Executable & 0.50 & 0.80 & 0.93 & 0.95 & 0.93 & 0.82 \\
\multirow{2}{*}{\texttt{PCI}} & No. Rules & 6{,}539 & 6{,}321 & 6{,}467 & 6{,}835 & 11{,}396 & 11{,}379 \\
 & \% Executable & 0.37 & 0.78 & 0.86 & 0.93 & 0.90 & 0.85 \\
\multirow{2}{*}{\texttt{Tax}} & No. Rules & 14{,}818 & 15{,}035 & 193 & 2{,}027 & 14{,}251 & 13{,}307 \\
 & \% Executable & 0.48 & 0.85 & 0.88 & 0.97 & 0.94 & 0.80 \\
\midrule
\multicolumn{8}{l}{\textbf{Dirty \& Type}} \\
\midrule
\multirow{2}{*}{\texttt{Beers}} & No. Rules & 1{,}977 & 2{,}071 & 2{,}267 & 2{,}398 & 2{,}484 & 2{,}475 \\
 & \% Executable & 0.32 & 0.85 & 1.00 & 1.00 & 0.98 & 0.95 \\
\multirow{2}{*}{\texttt{Flights}} & No. Rules & 1{,}637 & 1{,}694 & 1{,}965 & 2{,}015 & 2{,}548 & 2{,}836 \\
 & \% Executable & 0.43 & 0.82 & 0.99 & 1.00 & 0.97 & 0.90 \\
\multirow{2}{*}{\texttt{Hospital}} & No. Rules & 492 & 543 & 571 & 558 & 481 & 429 \\
 & \% Executable & 0.43 & 0.81 & 0.98 & 0.99 & 0.92 & 0.90 \\
\multirow{2}{*}{\texttt{PCI}} & No. Rules & 6{,}835 & 7{,}033 & 4{,}076 & 4{,}241 & 16{,}785 & 16{,}691 \\
 & \% Executable & 0.35 & 0.80 & 0.91 & 0.94 & 0.98 & 0.91 \\
\multirow{2}{*}{\texttt{Tax}} & No. Rules & 4{,}660 & 4{,}858 & 3{,}428 & 1{,}644 & 5{,}626 & 5{,}472 \\
 & \% Executable & 0.39 & 0.81 & 1.00 & 1.00 & 0.99 & 0.98 \\
\midrule
\multicolumn{8}{l}{\textbf{Just Dirty}} \\
\midrule
\multirow{2}{*}{\texttt{Beers}} & No. Rules & 2{,}451 & 2{,}466 & 4{,}052 & 4{,}115 & 2{,}686 & 2{,}602 \\
 & \% Executable & 0.34 & 0.76 & 0.68 & 0.83 & 0.89 & 0.58 \\
\multirow{2}{*}{\texttt{Flights}} & No. Rules & 2{,}503 & 2{,}494 & 4{,}819 & 4{,}892 & 8{,}581 & 8{,}033 \\
 & \% Executable & 0.46 & 0.78 & 0.98 & 0.99 & 0.97 & 0.92 \\
\multirow{2}{*}{\texttt{Hospital}} & No. Rules & 746 & 746 & 1{,}489 & 1{,}458 & 1{,}071 & 1{,}048 \\
 & \% Executable & 0.48 & 0.87 & 0.98 & 1.00 & 0.90 & 0.84 \\
\multirow{2}{*}{\texttt{PCI}} & No. Rules & 9{,}293 & 8{,}887 & 8{,}904 & 9{,}739 & 27{,}745 & 26{,}222 \\
 & \% Executable & 0.33 & 0.78 & 0.81 & 0.85 & 0.99 & 0.88 \\
\multirow{2}{*}{\texttt{Tax}} & No. Rules & 8{,}535 & 9{,}255 & 8{,}500 & 10{,}059 & 16{,}010 & 16{,}470 \\
 & \% Executable & 0.44 & 0.85 & 0.79 & 0.93 & 0.93 & 0.84 \\
\bottomrule
\end{tabular}
   \label{tab:step1_res}
 \end{table*}
 
\subsubsection{LLMs}
\label{subsubsec:llms}
To interface with the three selected local LLMs, we utilized the OpenAI Python client\footnote{https://github.com/openai/openai-python}. These models were selected based on their rankings in the Chatbot Arena LLM Leaderboard\footnote{https://lmarena.ai/?leaderboard} and their compatibility with our computing infrastructure:
\begin{itemize}
    \item \textbf{\glm{} (\glmLong{})\footnote{https://huggingface.co/cyankiwi/GLM-4.7-Flash-AWQ-4bit}:} a lightweight general-purpose model chosen for its ability to run on weaker hardware, serving as a lightweight baseline.
    \item \textbf{\qwen{} (\qwenLong{})\footnote{https://huggingface.co/cyankiwi/Qwen3-Coder-30B-A3B-Instruct-AWQ-4bit}:} a code-specialized model that may benefit the generation of syntactically correct rules.
    \item \textbf{\gemma{} (\gemmaLong{})\footnote{https://huggingface.co/RedHatAI/gemma-4-31B-it-FP8-Dynamic}:} a high-performance general-purpose model serving as a strong baseline for comparison.
\end{itemize}
This selection allows us to compare a lightweight model, a code-specialized model, and a high-capacity general-purpose model. 

\subsection{Evaluation of the Syntactic Filter}
\label{sec:eval-syntactic}
\Cref{tab:step1_res} shows the number of rules generated by the LLMs for each dataset, LLM, input variant, and $\mathcal{S}$ variant (i.e. with/without parameter). 
Interestingly, the models typically generated more than one rule per tuple. This contradicts our initial expectation of a single rule per tuple, since most tuples contain only one error affecting a single value. 
This suggests that the LLM generates additional rules to detect other potential errors, which may improve the generalizability of $\rulset$. However, this also increases the likelihood of overlapping and duplicated rules, further justifying the importance of the filtering pipeline.

 To evaluate the syntactic filter, we use the fraction $\frac{|\ro'|}{|\ro|}$ of syntactically correct rules as the metric. 
The findings in \Cref{tab:step1_res} indicate that the LLMs are reliable in generating executable rules, especially when $\mathcal{S}$ with parameter information is provided. Depending on the model and the dataset, 87.13\% to 100\%  of the rules are executable in that case.
This reliability was consistent across different datasets and various model, although the coding-specialized \qwen{} (mean executability = 91\%) and the general-purpose \gemma{} (mean executability = 93\%) outperformed the lightweight \glm{} (mean executability = 61\%).

\subsection{Evaluation of the Correctness Filter}
\label{sec:eval-correctness}

\begin{figure*}[ht]
    \centering
    \includegraphics[width=1\linewidth]{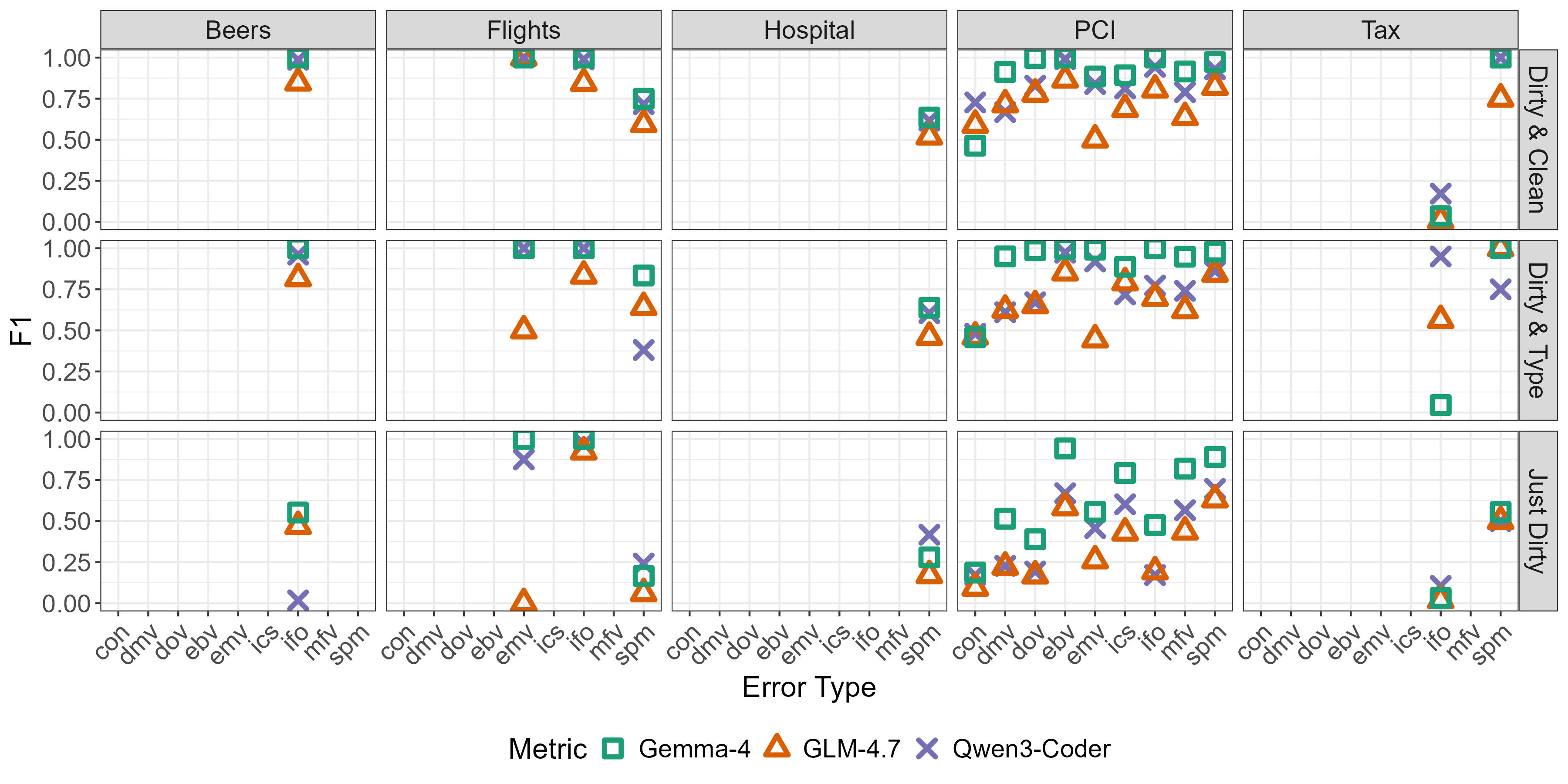}
    \caption{\StepTwo: Aggregated F1-Score per error type across the three LLMs, three input variants, and five datasets.}
    \label{fig:res_step2}
\end{figure*}

The correctness filter is applied to all rules that remain after the syntactic filter. As mentioned in \Cref{subsubsec:rule_verification}, we use F1 to evaluate whether the generated rules correctly identify the error in each tuple.

\Cref{fig:res_step2} presents the mean F1 per error type and per types of error information (\justDirty{}, \dirtyClass{} and \dirtyClean{}). Because the results of the rules generated with and without parameter information are very similar (max std: 0.087), we will present the mean results for both. Just Dirty performs worst overall, which is expected, as, in addition to creating rules, the dirty values have to be detected. Which is especially difficult cases where business knowledge is needed, for example for the \dov{} or the \ifo{} where the right format might depend of organizational preferences. The results for \dirtyClass{} and \dirtyClean{} demonstrate that it is possible to create successful rules for these error information types as well. The results for \dirtyClass{} indicate that providing information about the error type and column is enough to increase the overall performance by 0.33 compared to \justDirty{}.

Consistent with the observations from the \stepOne, \glm{} exhibited the lowest performance among the three models. Interestingly, \qwen{} also underperformed relative to \gemma{}. This is likely due to \gemma's better general reasoning capabilities, whereas the specialized coding capabilities that benefited \qwen{} in the \stepOne{} were less critical for ensuring the correctness of the rules.

Overall, the rule generation works well across all error types, with a mean F1 of 0.66, 
Performance varied by error type, with the LLM struggling most with \con{} (F1 = 0.4) and achieving the highest success with \ebv{} (F1 = 0.87). 
The success with \ebv{}, is likely due to these errors being easily detectable via regular expressions and only affecting a single value, given how they were generated. In contrast, \con{} involves inconsistencies across multiple columns, necessitating more complex rules.

\subsection{Evaluation of the Coverage Filter} 
\label{sec:eval-coverage}
\label{subsec:coverage_filter}

Following \Cref{subsubsec:coverage_filter}, we use recall as the metric since false positives have been eliminated in previous stages. A high recall indicates good generalizability across different variants of the same error type. To prioritize maximum coverage and avoid the risk of filtering out highly specific rules (as cautioned in \Cref{subsubsec:coverage_filter}), no minimum recall threshold was enforced during this stage. Consequently, all candidate rules were retained for the final redundancy filtering process.
 
\begin{figure*}[ht]
    \centering
    \includegraphics[width=1\linewidth]{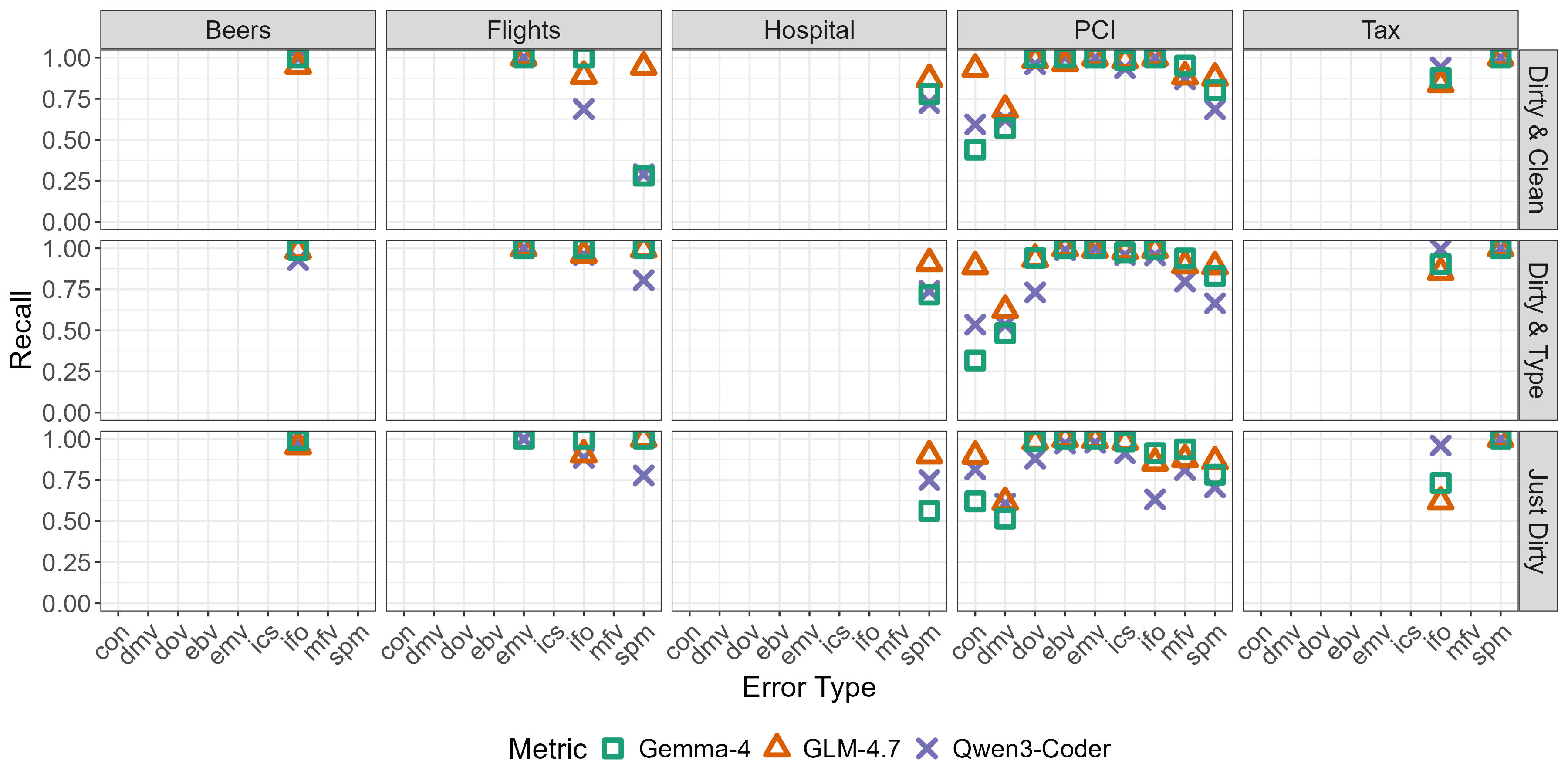}
    \caption{\StepThree: Aggregated recall per error type across the three LLMs, three input variants, and five datasets.}
    \label{fig:res_step3}
\end{figure*}
\Cref{fig:res_step3} shows that regardless of the specific type of error information provided, the rules demonstrate similar generalizability over other instances of the same type, achieving a mean recall of 0.88 (std = 0.16).
Contrary to the observations in the \stepOne{} and the \stepTwo{}, \glm{} is no longer the lowest-performing model in this stage. 
The rules generalize similarly across all LLM models (recall std = 0.16), although those generated by \qwen{} often perform the worst of the three. An exception is the \con{} error type, in which \gemma{} rules exhibit the lowest generalization. The reasons for this may be linked to the semantic complexity of \con{} errors, leading \gemma{} to produce overly specific rules to avoid specific contradictions.

Even though the generalization works for all error types, performance is notably lower for those with multiple variants, such as \dov{} (recall = 0.58), \con{} (recall = 0.67), and \spm{} (recall = 0.84).
For example, for the \city{} column, the LLM created a rule with a regex that correctly detects values starting with a lowercase letter. However, the rule does not generalize to all Austrian city names: cities containing umlauts (e.g., ``ö'', ``ü'', ``ä''), hyphens, or multiple words (e.g., ``Hagenberg im Mühlkreis'') are not detected, making the rule overly general but not optimal.

\begin{figure*}[ht]
    \centering
    \includegraphics[width=1\linewidth]{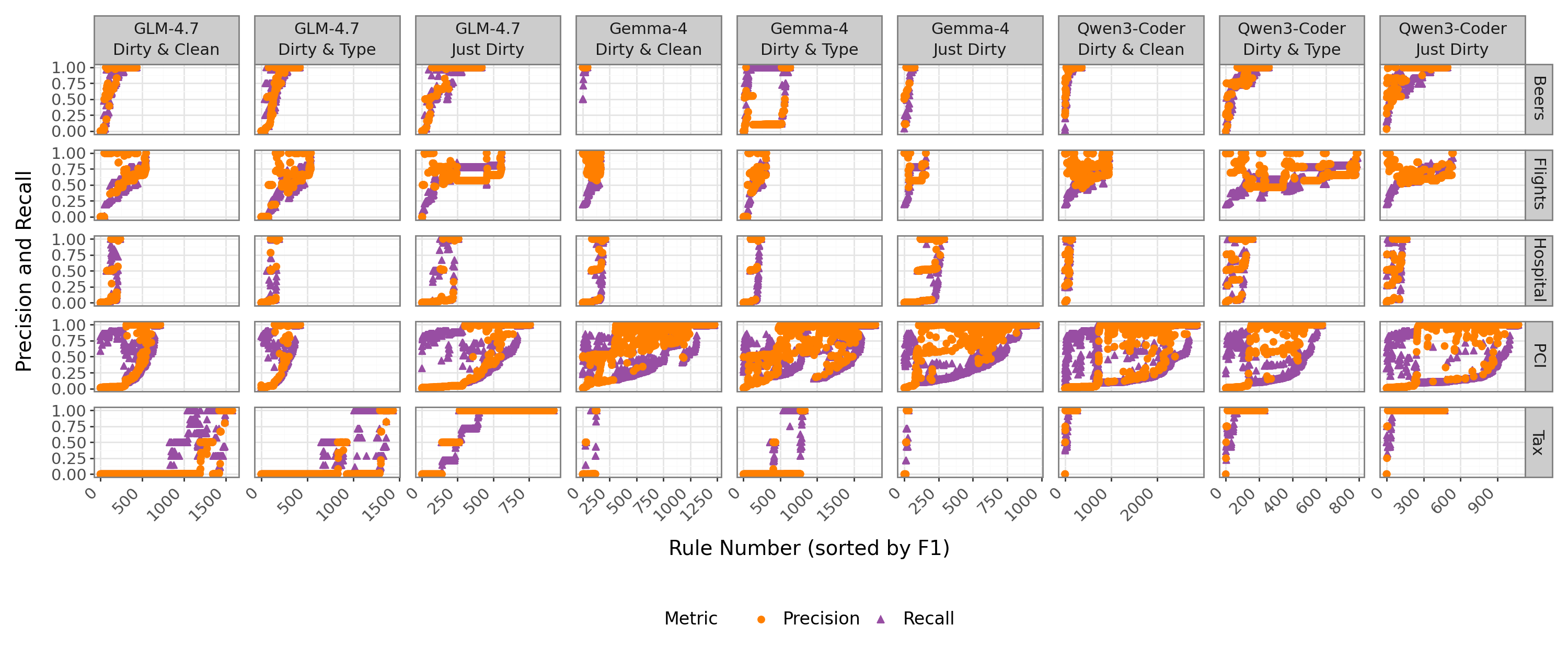}
    \caption{Precision and recall of each rule measured on an unseen dataset $\dtest$ for each of the three LLMs, three data input variants, and five datasets. The results have been ordered by F1 score.}
    \label{fig:step4_metric}
\end{figure*}

\begin{figure*}
    \centering
    \includegraphics[width=1\linewidth]{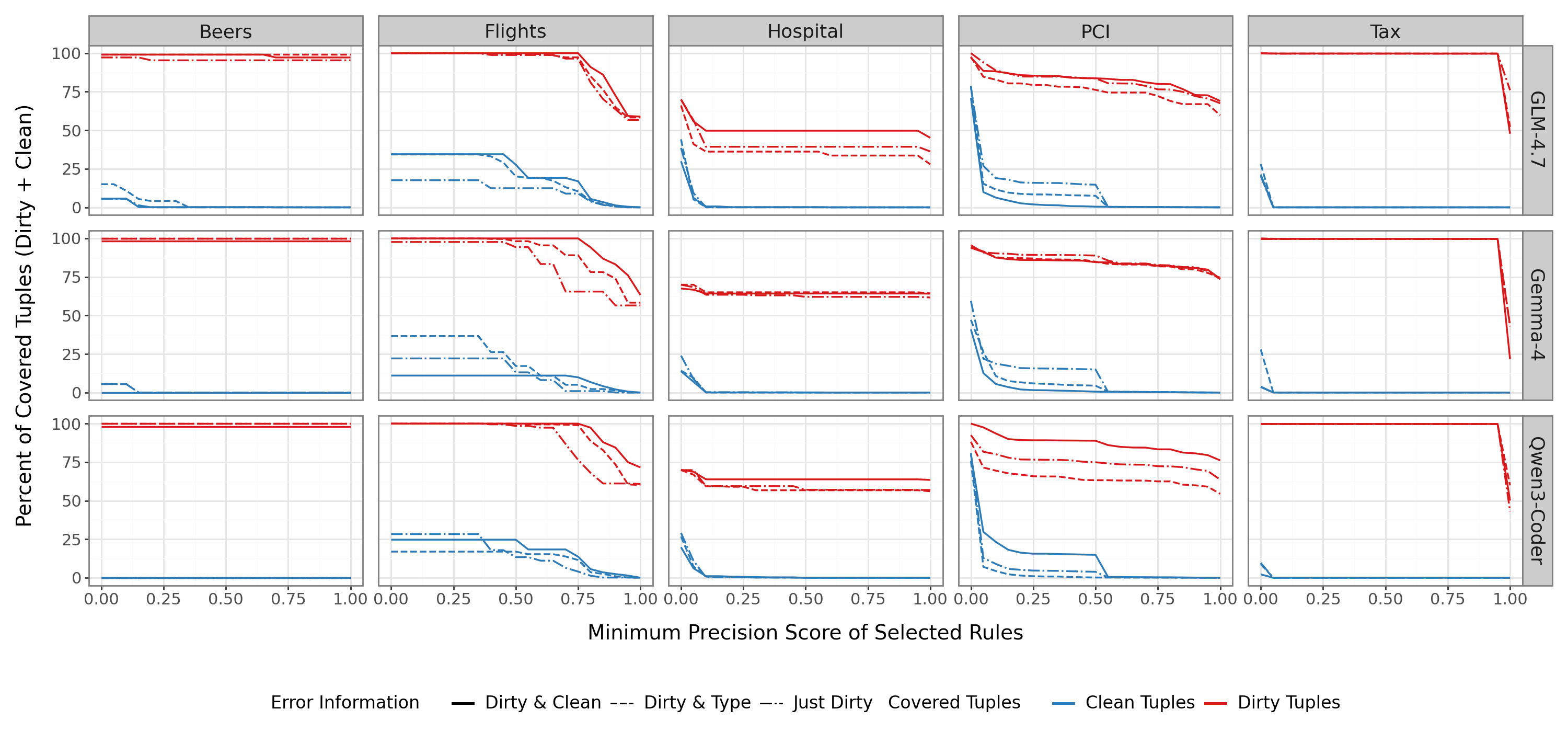}
\caption{\StepFour: Percentage of covered clean and dirty tuples found by the rules in $\rulset$, calculated for each of the five $\dtest$, three input variants and three LLMs with different precision settings for the rules in $\ro'''$.}
    \label{fig:step4_coverage}
\end{figure*}
\subsection{Evaluation of the Redundancy Filter}
\label{sec:eval-redundancy}
Before applying the redundancy filter as described in \Cref{subsubsec:redundancy_filter}, we evaluate how well each rule in $\ro'''$ generalizes to a new, unseen dataset $\dtest$. For this, recall and precision are computed per rule; see \Cref{fig:step4_metric} for the results. The figure shows that the rules in $\ro'''$ generalize well to unseen data, as most of the rules either have a high precision or high recall and a subset exhibits both high precision and high recall.
These results are consistent with those reported in the \stepThree{}, which already demonstrated that the LLM is capable of generating rules that generalize to unseen data but may miss some dirty tuples. Taken together, the individual rule quality suggests that the rule set as a whole should perform well in combination.

We then apply the redundancy filter to $\ro'''$ and evaluate the resulting final rule set $\rulset$ against $\dtest$. For the \texttt{beers}, \texttt{flights}, \texttt{hospital}, and \texttt{tax} dataset, we aggregate the results over 5 folds, as only one dataset is available. For the \texttt{PCI} we generated a $\dtest$ with 1,000,000 tuples 25\% of them dirty ones. \Cref{fig:step4_coverage} shows the total number of detected tuples split into TP (dirty tuple) and FP (clean tuple) across different minimum precision thresholds. At a precision of 1 (i.e., no false positives), the rules achieve at least 50\% coverage (except for the \texttt{tax} dataset), depending on the error type existing in each dataset.
For the \texttt{tax} dataset, a slight increase in false positives (up to 6 FP or 0.007\%) results in near-complete error coverage (99.58\%), indicating a favorable trade-off. In contrast, for the \texttt{flights} dataset, relaxing the precision threshold yields comparatively smaller gains of 9\% in coverage while introducing a larger number of false positives (+6.7\%), suggesting a more conservative threshold selection. This supports our design choice to filter and rank rules by precision, allowing data stewards to adapt the rule set to different error characteristics and tolerance levels.
Furthermore, the redundancy filter significantly reduces the number of rules, resulting in the final rule set $\rulset$ only containing between 2 and 136 rules. Lowering the precision threshold increases coverage at the cost of introducing false positives, exposing a controllable precision–coverage trade-off.
This trade-off remains consistent across the different LLMs for the \texttt{beers}, \texttt{flights}, and \texttt{tax} datasets. However, for the \texttt{hospital} and \texttt{PCI} datasets, \gemma{} exhibits the most favorable trade-off, followed closely by \glm{} and \qwen{}.

\subsection{Summary of Results}
Our experiments show that an LLM is able to generate syntactically correct rules with executability rates ranging from 87.13\% to 100\% when provided with both rule names and valid parameters. With purposeful filtering, those rules can be reduced to a small rule set that generalizes well to unseen data. 

The results indicate that even providing only the error type and column increases the overall performance (F1 score) by an average of 0.33 compared to providing only dirty values. Generally speaking, the rule generation works well across most error types, yielding a mean F1 score of 0.66, although complex error types such as contradictions showed lower performance (F1 = 0.4). The generalizability of the generated rules is lower for those error types with multiple variants, such as \dmv{} (recall=0.58) or \con{} (recall=0.67), compared to the mean recall of 0.87 across all types.

The evaluation of the redundancy filter demonstrates that at a precision of 1 the reduced rule set achieves at least 50\% coverage of the dirty tuples across four of the five unseen datasets. 
Decreasing this precision threshold can, in some cases, improve coverage significantly. For example, in the \texttt{tax} dataset, introducing only 6 false positives increased the percentage of found dirty values by approximately 70 percentage points to reach near-complete coverage. This highlights a controllable trade-off between precision and recall that a data steward can adjust based on the tolerance for false positives in the respective organization and use case.

\section{Related Work}
\label{sec:related-work}
We discuss related work along three lines: rule learning in the machine learning (ML) community, DQ rule discovery in the database community, and LLM-based approaches to DQ rule generation.

\paragraph{Rule learning in the ML community}
Learning interpretable error-correcting rules has a long tradition in machine learning in general, and in inductive rule learning in particular. Generalized additive models~\cite{GAMs} may be viewed as a family of algorithms that learn multiple layers classifiers where each layer incrementally improves upon the previous ones. In particular boosting-based approaches~\cite{AdaBoost} have a clear focus on learning models that correct classification errors. Patching~\cite{sk:AAAI-18} takes a somewhat different angle in that it investigates the potential of learning interpretable rules that characterize regions of errors in an underlying immutable black-box system. \citet{CharacterizingErrors-NLP} propose a system for learning rules to correct an underlying NLP system. 
Collectively, these approaches provide important foundations for learning human-understandable rule sets from examples and focus on improving predictive systems by characterizing or correcting model errors.
In contrast, \fname{} focuses on generating DQ rules that are executable by a DQ tool to detect errors in structured datasets.

\paragraph{DQ rule discovery in the database community}
Within the database community, automatic discovery of DQ rules has been studied primarily in the context of integrity constraints and functional dependencies. 
\citet{Chiang_2008} and \citet{Yeh_2010} explored the discovery of DQ rules from data, focusing on conditional functional dependencies and related dependency patterns. These traditional approaches are largely combinatorial, and rely on discovering statistical regularities across large datasets. 
As a result, they are well suited for structural constraints, but less effective to learn semantic validation rules that depend on contextual information or domain knowledge. 
To overcome these limitations, recent work explores LLMs for generating higher-level semantic constraints.

\paragraph{LLM-based DQ rule generation}
The use of LLMs to generate interpretable rules and features has received increasing attention.
\citet{LLMs-FeatureGeneration} propose LLMs for generating tabulated features for natural language documents, and \citet{Chat2Data-Gottlob} propose the use of LLMs for automatic database curation via the Chat2Data framework. 

\begin{table}[htbp]
\centering
\caption{Comparison of \fname{} with related LLM-based rule generation frameworks. Xie et al.~\cite{Xie_DQRLLMmed_2025} and Schneider~\cite{Schneider_LLMBagOfRules_2025}.}
\label{tab:comparison}
\begin{threeparttable}
\begin{tabular}{lccc}
\toprule
\textbf{Step} &  \textbf{Xie et al.}
&\textbf{Schneider}& \textbf{\fname} \\ 
\midrule
Input &  \makecell{Requirements +\\Metadata} 
&\makecell{Template +\\Metadata} & Dirty tuples \\
Generation &  One-time &One-time & Incremental \\
\midrule
 \multicolumn{4}{l}{\textit{Filtering}}\\
\quad Syntactic &  \cmark
&\cmark\tnote{a}& \cmark \\
\quad Correctness &  \xmark 
&\cmark\tnote{a} & \cmark \\
\quad Coverage &  \xmark
&\xmark & \cmark \\
\quad Redundancy &  \cmark 
&\cmark\tnote{a}& \cmark \\
\midrule
Output &  \makecell{SQL\\rules}&\makecell{Prolog-like\\rules}& \makecell{GX\\rules} \\
\bottomrule
\end{tabular}
\begin{tablenotes}
\footnotesize
\item[a] Requires / involves a LLM-as-a-judge. 
\end{tablenotes}
\end{threeparttable}
\end{table}

Most closely related to our work are two approaches that also use LLMs to generate DQ rules, which we compare along the lines of our generate-filter formalization in \Cref{tab:comparison}. \citet{Xie_DQRLLMmed_2025} follow a requirement-driven approach in which rules are derived from natural-language business requirements and database metadata, tailored to the medical domain of electronic health records. \citet{Schneider_LLMBagOfRules_2025} proposes a template-driven approach where templates specify the tables and operations an LLM should use to generate the rules. 
Both approaches generate a complete rule set from scratch in a one-time manner, based on what errors the LLM expects rather than on errors observed in the data.
In contrast, \fname{} generates rules from observed (i.e., real) errors in the data: for each dirty tuple, \fname{} (1)~detects the respective error, (2)~suggests a rule to detect this error, and (3)~applies the filtering steps outlined in \Cref{sec:approach-ledger}. Rather than generating rules from scratch, \fname{} iteratively refines an existing rule base from a small number of examples, relieving data stewards from manually specifying rules or requirements~\cite{Schneider_LLMBagOfRules_2025}. 
Consequently, a direct quantitative comparison with \citet{Xie_DQRLLMmed_2025} and \citet{Schneider_LLMBagOfRules_2025} is not meaningful, since the three approaches differ in their required input as well as their output (i.e., rule syntax). While \citet{Xie_DQRLLMmed_2025} require domain-specific business requirements and \citet{Schneider_LLMBagOfRules_2025} require manually specified requirements, \fname{} can be used solely on dirty tuples as input, which reflects a more practical scenario.

\section{Conclusion}
\label{sec:conclusion}

In this work, we introduced \fname{}, an LLM-based generate-filter framework that generates DQ rules for the enterprise DQ tool \Gx{}. Unlike previous works~\cite{Xie_DQRLLMmed_2025,Schneider_LLMBagOfRules_2025}, which rely on the manual specification of domain-specific requirements or templates, \fname{} targets a practical setting: a data steward has identified a dirty tuple that passed the validation unnoticed and needs to extend the rule set of \Gx{} accordingly. In the \emph{generate} step, an LLM generates candidate rules from the dirty tuple. In the \emph{filter} step, four filters remove low-quality rules. They check the executability and correctness of individual rules, their generalizability to other errors of the same type, and remove redundant rules from the final rule set.

Our evaluation across five datasets demonstrates that \fname{} generates correct and generalizable rules from dirty tuples alone. Providing additional information about the error, such as its type and location, or the clean tuple, further improves the rule quality. Importantly for real-world scenarios, the error type and location alone are sufficient to significantly improve performance. This reduces the need to find clean data for the successful generation of DQ rules. For less complex error types, the dirty tuple alone can still yield sufficient rule sets. Furthermore, \fname{} performs consistently across different LLMs. Interestingly, higher-capacity general-purpose models outperform lightweight or coding-specialized models as error complexity increases. This suggests that general reasoning is more critical for rule correctness than specialized coding proficiency. Finally, we observed an occasional inverse relationship between initial rule correctness and generalizability. This highlights the importance of the generate-filter approach, which filters a wide range of candidate rules.

\textbf{Future work.}
Based on our industry partner's feedback, we plan to systematically evaluate the ability of LLMs to generate natural-language descriptions of the generated rules. Such descriptions also allow non-technical users to understand and use these rules and automate documentation processes. 
We consider this direction promising since our manual reviews revealed that the LLM occasionally produced useful rule descriptions without explicit prompting.
Furthermore, we aim to explore methods for automatically detecting the error type of a given dirty tuple. Our results already demonstrate its potential: providing the error type significantly improves the rule generation performance.

\begin{acks}
The research was supported by the Austrian ministries BMIMI, BMWET and the State of Upper Austria in the frame of the SCCH COMET competence center INTEGRATE (FFG 892418) and by the “ICT of the Future” project QuanTD (no. 898626), as well as by the Austrian Science Fund FWF (10.55776/COE12).
RedHatAI/gemma-4-31B-it-FP8-Dynamic and Claude Pro were used to generate code and code documentation for the experimental evaluation, and to revise and proofread parts of the paper. All ideas and the scientific content of this paper originate from the authors. Where text was AI-assisted, it was  rewritten and verified by the authors, who take full responsibility for the content of this work.
\end{acks}

\section*{Artifacts}
The code implementing \fname{} is available in a public GitHub repository\footnote{https://github.com/Anna-Christina-Glock/dq\_rule\_generation}. The code to generate the \texttt{PCI} dataset is also available in a public GitHub repository\footnote{https://github.com/Anna-Christina-Glock/pci-llm-toolkit/tree/main/data-generator}, and the other four (polluted and labeled) datasets are also publicly available\footnote{https://github.com/Anna-Christina-Glock/dq\_rule\_generation}. The LLMs used during the experiments are available via HuggingFace\footnote{https://huggingface.co/models}.

\bibliographystyle{ACM-Reference-Format}
\bibliography{references}

\end{document}